\documentclass[10pt,twocolumn,letterpaper]{article}

\usepackage[pagenumbers]{wacv} 

\usepackage{graphicx}
\usepackage{amsmath,amssymb,amsfonts}
\usepackage{booktabs}
\usepackage{algorithm}
\usepackage{algorithmic}
\usepackage{pgfplots}
\usepackage{multirow}
\usepackage{layouts}
\usepackage{array}
\usepackage{colortbl}
\usepackage{xcolor}
\usepackage{amsmath}

\pgfplotsset{width=10cm,compat=newest}

\usepgfplotslibrary{external}
\DeclareMathOperator*{\argmax}{argmax} 
\usepackage[pagebackref,breaklinks,colorlinks]{hyperref}

\usepackage[capitalize]{cleveref}
\crefname{section}{Sec.}{Secs.}
\Crefname{section}{Section}{Sections}
\Crefname{table}{Table}{Tables}
\crefname{table}{Tab.}{Tabs.}

\def\wacvPaperID{1089} 
\def\confName{WACV}
\def\confYear{2025}

\begin{document}

\title{When to Extract ReID Features: A Selective Approach for Improved Multiple Object Tracking}

\author{Emirhan Bayar\\
Middle East Technical University\\
Ankara, Turkey\\
{\tt\small emirhan.bayar@metu.edu.tr}
\and
Cemal Aker\\
Middle East Technical University\\
Ankara, Turkey\\
}

\vspace{-1cm}

\maketitle

\begin{abstract}

Extracting and matching Re-Identification (ReID) features is used by many state-of-the-art (SOTA) Multiple Object Tracking (MOT) methods, particularly effective against frequent and long-term occlusions. While end-to-end object detection and tracking have been the main focus of recent research, they have yet to outperform traditional methods in benchmarks like MOT17 and MOT20. Thus, from an application standpoint, methods with separate detection and embedding remain the best option for accuracy, modularity, and ease of implementation, though they are impractical for edge devices due to the overhead involved. In this paper, we investigate a selective approach to minimize the overhead of feature extraction while preserving accuracy, modularity, and ease of implementation. This approach can be integrated into various SOTA methods. We demonstrate its effectiveness by applying it to StrongSORT and Deep OC-SORT. Experiments on MOT17, MOT20, and DanceTrack datasets show that our mechanism retains the advantages of feature extraction during occlusions while significantly reducing runtime. Additionally, it improves accuracy by preventing confusion in the feature-matching stage, particularly in cases of deformation and appearance similarity, which are common in DanceTrack.
\href{https://github.com/emirhanbayar/Fast-StrongSORT}{https://github.com/emirhanbayar/Fast-StrongSORT}
\href{https://github.com/emirhanbayar/Fast-Deep-OC-SORT}{https://github.com/emirhanbayar/Fast-Deep-OC-SORT}

\end{abstract}

\section{Introduction}
\label{sec:intro}

\begin{figure}
    \centering
    \includegraphics[width=0.5\textwidth]{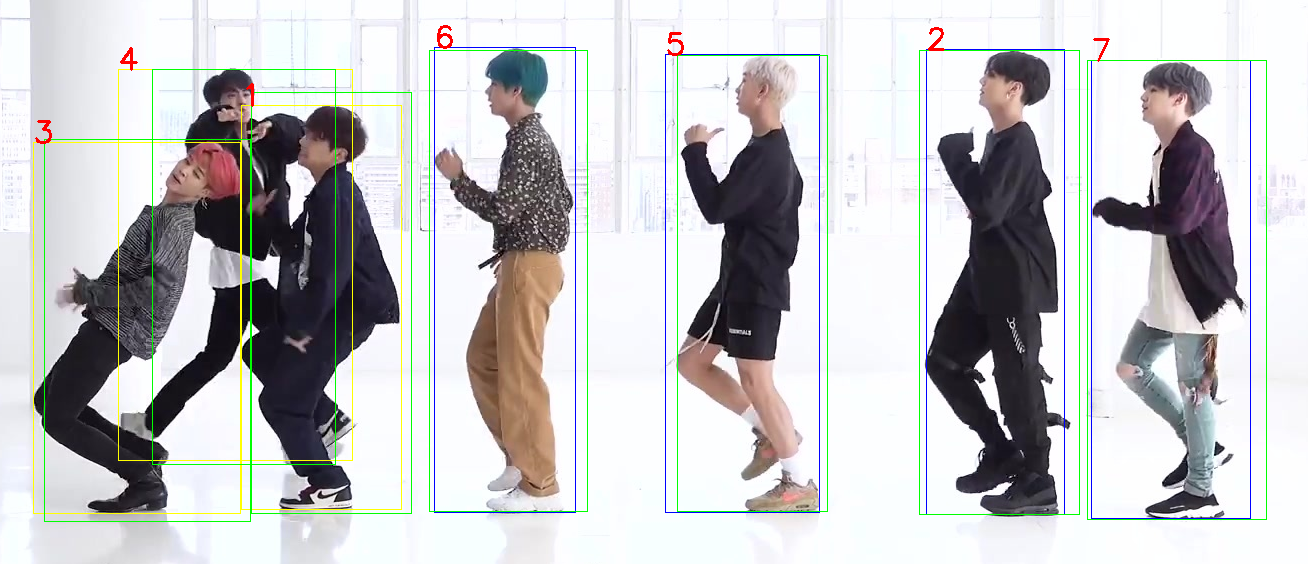}
    \caption{In this scene, the \textcolor{green}{green boxes} are the bounding boxes of the tracklets. The \textcolor{blue}{blue boxes} are the detections with no occlusion risk. The \textcolor{yellow}{yellow boxes} are the detections with occlusion risk. The starting point of the proposed mechanism is: Why do we need to extract features for the blue boxes while matching them via positional information is not risky?}
    \label{fig:example_scene}
\end{figure}

Multiple Object Tracking (MOT) is the problem of maintaining the identity of objects of predefined classes across a sequence of frames \cite{strongsort}. Tracking-by-detection (TbD) has long been the dominant paradigm in the field. Methods that follow this paradigm divide tracking into two steps: (1) frame-wise object detection, and (2) data association to link those detections to form trajectories \cite{Seidenschwarz_2023_CVPR}. Real-time TbD methods commonly hold a Kalman Filter \cite{kalman} for each tracked object, match the estimated states of the objects with the detections via the Hungarian algorithm \cite{hungarian} and update the states of the tracklets with matched detection\cite{sort, deepsort, bytetrack, ocsort, deepocsort,strongsort,botsort,boostTrack,pia2,imprasso,lgtrack}.The cost matrix of the matching algorithms can be based on the object's position, motion, appearance, or a combination of multiple factors \cite{deepsort, ocsort, deepocsort}.

ReID features \cite{reid-bot, fastreid} are widely used in the matching phase of state-of-the-art (SOTA) methods due to their robustness against occlusions and clutter \cite{deepocsort, deepsort, strongsort, botsort, fairmot, rethink}. These methods, which rely on computationally expensive Convolutional Neural Networks (CNNs) for feature extraction, result in extremely low FPS (Frames per Second) when executed on resource-constrained hardware \cite{runtime}. On the other hand, existing MOT methods prioritize achieving high accuracy on prerecorded datasets such as MOT17 \cite{mot16-review}, MOT20 \cite{mot20}, and DanceTrack \cite{dancetrack}. However, in real-time image streaming, frames are captured at wider intervals as FPS decreases, making changes in object positions and movements more significant. Consequently, the error in prediction and matching steps increases \cite{zhou2022apptracker, fpsanalysis}. This renders the mentioned methods impractical for edge artificial intelligence applications.

Several attempts have been proposed to mitigate the overhead of feature extraction resulting from separate detection and tracking. The idea of joint detection and embedding (JDE) \cite{jde-review} is one such approach, which enables detection and feature extraction in a shared backbone \cite{jde1, jde-review, rethink, retinatrack,fairmot}. Some other approaches replaces data association by regression of previous detections \cite{tbreg1, tbreg2}. With the introduction of DETR \cite{detr}, using transformers, which learn spatiotemporal relation of detections and tracklets, for end-to-end object detection and tracking gained popularity \cite{trans1, trans2, trans3, trans4}. However, they are criticized for lack of modularity and computational demand of training and deployment \cite{Seidenschwarz_2023_CVPR,moyolo}.

 While end-to-end object detection and tracking are trending and intriguing from a research perspective, they have yet to outperform traditional methods in well-known benchmarks like MOT17 and MOT20. Therefore, TbD methods with separate detection and embedding remain the best option from an application point of view due to their accuracy, modularity, and ease of implementation \cite{boxmot, ocmc, conftrack}. Instead of eliminating the feature extraction step, a recent study \cite{ss*} proposed an elegant strategy to minimize the number of detections on which feature extraction is performed. Although this idea is inspirational and outperforms trackers that do not use ReID features, it does reduce the accuracy compared to the original tracker on which the algorithm is applied. In this study, we propose extensions to this algorithm to limit the number of extractions without sacrificing accuracy—in fact, increasing it. The contributions of this work can be summarized as follows:

 \begin{itemize}
     \item A selective mechanism to minimize the overhead of feature extractions while preserving accuracy, modularity, and ease of implementation. This method can be applied to any real-time TbD tracker that uses a state estimator (e.g. Kalman Filter) and performs matching based on ReID features.

    \item A feature decay mechanism that increases accuracy by enhancing the temporal locality of matched features.
     
    \item A showcase of its implementation on StrongSORT and Deep OC-SORT, with detailed analysis on MOT17, MOT20, and DanceTrack datasets and runtime analysis on an edge device.
 \end{itemize}

\section{Realted Work}
\label{sec:related-work}

\subsection{Tracking-by-Detection (TbD)}
\label{subsec:tbd}

The methods that follow the TbD paradigm have two essential components: object detection and object association \cite{mot16-review, review}. In these methods, the update of the tracklets solely depends on the matched detections. This makes the object detection step indispensable. On the other hand, the object association step is responsible for determining the correspondence between the detections and the tracklets. The methods are distinguished by the way they perform the object association step.

SORT \cite{sort} is a baseline method that uses the Kalman filter to predict the position of the object in the next frame and the Hungarian algorithm to associate the detections with the tracklets. The cost matrix for the Hungarian algorithm is computed based on the intersection over union (IoU) of the detections and the tracklets. However, the IoU cost by itself is not enough for robustness to the occlusions and misdetections.

DeepSORT \cite{deepsort} is an extension of SORT that uses a deep neural network to extract the appearance features of the detections and use them in the cost matrix of the matching phase. Because of its robustness against frequent and long-term occlusions, feature extraction has become a key component for the methods that prioritize accuracy over speed. Different combinations of cost matrix construction, feature update/bank mechanisms, and matching strategies have been proposed \cite{botsort, strongsort, deepocsort, bytetrack}.  On the other hand, some methods that prioritize speed over accuracy \cite{stadler-2021,stadler-2022,diffmot} ignore the feature extraction step and make use of the motion information to associate the detections with the tracklets. However, it has been criticized that these methods are making assumptions about the movement patterns of the objects \cite{smiletrack}, and may not work well in crowded scenes.

\subsection{Joint Detection and Embedding (JDE)}
\label{subsec:jde}

To eliminate the overhead of feature extraction, the concept of joint detection and embedding (JDE) \cite{jde-review} was introduced, integrating detection and feature extraction into a single unified backbone. Tong et al. \cite{jde0} added a fully connected branch to Faster R-CNN \cite{fasterrcnn} to obtain ReID features, and fused an instance-based classification loss in addition to classification and regression losses. JDE \cite{jde-review} method replaced Faster R-CNN which is a two-stage detector with Yolov3 \cite{yolov3} and instance-based classification with triplet loss \cite{triplet}. FairMOT \cite{fairmot} tackled the problems caused by anchors and trained the CenterNet \cite{centernet}, which is an anchor-free detector, by fusing an instance-based classification loss. The major problem with the notion of JDE is the competition between re-identification and object classification tasks. While the former task needs features that discriminate instances of the same class, the latter needs common features that all instances of the same class possess and are discriminative against other classes. CSTrack \cite{rethink} attempted to tackle this by decoupling extracted features to be fed into object classification and re-identification heads through separate channels. While CSTrack improves association-related metrics, it falls behind in higher-order metrics such as HOTA \cite{hota}, particularly under conditions like blur and over-occlusion that degrade detection performance.

\subsection{Joint Detection and Tracking without ReID}
\label{subsec:jdt}

Joint detection and tracking, or end-to-end tracking, aims to eliminate ReID feature matching by proposing a data-driven association of motion priors. Chained-tracker \cite{ctracker}, Tracktor \cite{tracktor}, and TubeTK \cite{tubetk} regress detections across frames to generate trajectories. Recently, transformer-based end-to-end models gained popularity for their ability to model contextual and spatiotemporal information \cite{trans1, trans2, trans3, trans4}. Despite their success on DanceTrack, they fall short in MOT17 and MOT20 benchmarks. Moreover, they are known for their complex training strategies and prolonged training times \cite{Seidenschwarz_2023_CVPR, latesttrans}. Additionally, their inference speed may not surpass the methods with separate detection and tracking \cite{smiletrack}, which are top performers in MOT17 and MOT20 benchmarks and are easy to implement and extend thanks to their modularity.

\subsection{Selective Feature Extraction}
\label{subsec:sfe}

Recently, Bayar et al. proposed an algorithm that decouples detections according to the possible number of candidate tracklets \cite{ss*}. By a simple Intersection over Area (IoU) check, the detections with a single possible candidate are gated from feature extraction and matching, while ReID features are extracted from those with no or multiple candidates. Despite its elegance in minimizing the overhead of feature extraction while preserving the modularity and keeping the accuracy way above the trackers without ReID matching, it reduces the accuracy compared to the original method to which the algorithm was applied.

\section{Proposed Method}


Our method can be applied to any existing TbD tracker that uses Kalman Filter. These trackers typically match detections with the existing tracklets for each frame. In that manner, tracklets can be considered as candidates for detections. Our baseline will be the mechanism as proposed in \cite{ss*}, which simply detects detections with a single candidate tracklet and exempts them from feature extraction. In this section, we first examine the base mechanism and its drawbacks, and then introduce new modules to avoid them.

\subsection{Base Mechanism}

\begin{table*}[ht]
\centering
\caption{Performance comparison of StrongSORT with FSS on MOT17 and DanceTrack validation sets.}
\label{tab:runtime}
\resizebox{0.9\textwidth}{!}{
\begin{tabular}{lccccccc}
\toprule
Tracker & PDE* $\downarrow$ & FET** (s) $\downarrow$ & FPS $\uparrow$ & HOTA $\uparrow$ & AssA $\uparrow$ & IDF1 $\uparrow$ \\
\midrule
\multicolumn{7}{c}{MOT17 Val} \\
\midrule
StrongSORT & 100.0 & 2457 & 0.96 & 69.54 & 73.29 & 82.17 \\
FSS 0.0 & 76.38 (-24\%) & 1946 (-21\%) & 1.17 (22\%) & 69.36 (-0.26\%)& 72.92 (-0.50\%) & 81.88 (-0.35\%) \\
FSS 0.1 & 58.67 (-41\%) & 1598 (-35\%) & 1.40 (46\%) & 69.39 (-0.22\%)& 72.93 (-0.49\%) & 81.81 (-0.44\%) \\
FSS 0.2 & 43.93 (-56\%) & 1241 (-50\%) & 1.73 (80\%) & \textcolor{green}{\textbf{69.61 (0.1\%)}}& \textcolor{green}{\textbf{73.44 (0.21\%)}} & \textcolor{green}{\textbf{82.56 (0.47\%)}} \\
FSS 0.3 & 33.40 (-67\%) & 995 \phantom{1}(-60\%) & 2.07 (116\%) & 69.09 (-0.65\%)& 72.50 (-1.07\%) & 81.97 (-0.24\%) \\
FSS 0.4 & 23.46 (-77\%) & 772 \phantom{1}(-69\%) & 2.52 (163\%) & 68.31 (-1.77\%)& 70.87 (-3.30\%) & 80.91 (-1.53\%) \\
FSS 0.5 & 16.14 (-84\%) & 547 \phantom{1}(-78\%) & 3.21 (234\%) & 68.86 (-0.98\%)& 71.88 (-1.92\%) & 81.21 (-1.17\%) \\
\midrule
\multicolumn{7}{c}{DanceTrack Val} \\
\midrule
StrongSORT & 100.0 & 14054 & 1.56 & 56.61 & 41.12 & 55.91 \\
FSS 0.0 & 85.34 (-15\%) & 11972 (-15\%) & 1.79 (15\%) & \textcolor{green}{\textbf{57.07 (0.81\%)}} & \textcolor{green}{\textbf{41.78 (1.61\%)}} & \textcolor{green}{\textbf{56.35 (0.77\%)}} \\
FSS 0.1 & 73.40 (-27\%) & 10518 (-25\%) & 2.00 (28\%) & 56.55 (-0.11\%) & 41.01 (-0.27\%) & 55.90 (-0.03\%) \\
FSS 0.2 & 58.64 (-41\%) & 8548 \phantom{1}(-39\%) & 2.38 (53\%) & \textcolor{green}{\textbf{56.80 (0.34\%)}} & \textcolor{green}{\textbf{41.36 (0.60\%)}} & \textcolor{green}{\textbf{56.19 (0.49\%)}} \\
FSS 0.3 & 43.02 (-57\%) & 6434 \phantom{1}(-54\%) & 2.97 (90\%) & 55.46 (-2.03\%) & 39.38 (-4.22\%) & 54.06 (-3.32\%) \\
FSS 0.4 & 27.97 (-72\%) & 4264 \phantom{1}(-70\%) & 4.02 (158\%) & 53.92 (-4.75\%) & 37.29 (-9.32\%) & 52.89 (-5.40\%) \\
FSS 0.5 & 16.28 (-84\%) & 2540 \phantom{1}(-82\%) & 5.64 (262\%) & 51.55 (-8.94\%)  & 34.17 (-16.90\%) & 50.92 (-8.93\%) \\
\bottomrule
\multicolumn{4}{c}{*PDE is the abbreviation of Percentage of Detections with Extraction.} \\
\multicolumn{3}{c}{**FET is the abbreviation of Feature Extraction Time.} \\
\end{tabular}
}
\end{table*}

Given a set of detections, for each detection, the base mechanism calculates the IoU between the particular detection and all the confirmed tracklets.  Confirmed tracklets are the ones that are matched with detection at least once in the previous frames. If exactly one confirmed tracklet exists with IoU greater than a certain threshold, the detection is marked non-risky to match without appearance features; otherwise, it is marked as risky. If detection is marked as non-risky, the tracklet with the largest IoU with the detection is considered as a candidate as described in equation \ref{eq:IoU} where ct(d) denotes candidate to detection d. This method was applied on StrongSORT \cite{strongsort}, whose matching is two-staged, first depending on sole appearance features and second depending on sole IoU cost. The non-risky detections are directly assigned with the infinite appearance feature vector to have infinite cosine distance to all the tracklets, and so exempted from the first stage of the matching to be matched in the second stage because of having only one candidate.


\begin{equation}
    \label{eq:IoU}
    \text{ct}(d) = 
    \begin{cases}
        \argmax\limits_{t \in T} (\text{IoU}(d, t)) & \text{if } |\{t \in T | \text{IoU}(d, t) > \theta_{\text{IoU}}\}| = 1 \\
        \emptyset & \text{otherwise}
    \end{cases}
\end{equation}

This mechanism has three main drawbacks:

\begin{itemize}
    \item Gating the detections from the first stage of matching may be unsafe. Additionally, the method assumes the first stage will solely depend on the appearance feature, and it will not be applicable otherwise.

    \item In case of missing detections. The tracklet that is supposed to be matched with the missing detection can be considered as a candidate for another detection. Additionally, one tracklet can be marked as a candidate for more than two detections misleadingly. 
    
    \item By avoiding feature extraction, the mechanism disrupts the feature update \cite{strongsort, deepocsort, botsort} mechanisms that are used to preserve long-term information.

\end{itemize}

\subsection{Ensure the Match Instead of Gating}

Instead of being given an infinite vector as in the base mechanism, the appearance features of non-risky detections are directly copied from their candidates. In other words, the contribution of the appearance features is set to zero during matching if there is only one candidate tracklet for detection. Considering some of the existing trackers \cite{strongsort,botsort} sets the cost of the appearance features to maximum if the overlap is less than a certain threshold, this method does it the other way around. It sets the cost of the appearance features to zero if there exists only one tracklet with a high IoU. This way, they will be matched because of zero feature cost and a single candidate depending on motion cue.

\subsection{Aspect Ratio Similarity}
\label{subsec:aspect_ratio_similarity}

\begin{figure}
    \centering
    \resizebox{0.43\textwidth}{!}{
    \input{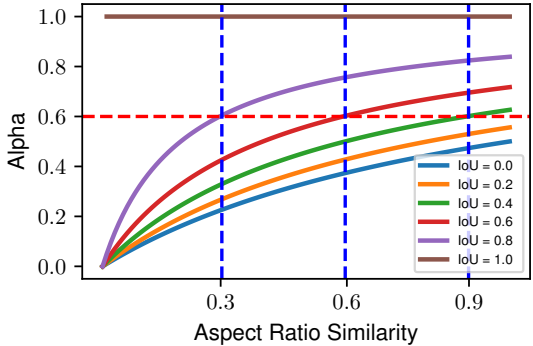}
    }
    \caption{Aspect ratio similarity vs $\alpha$ for different IoU values.}
    \label{fig:alphavsV}
\end{figure}

\begin{figure*}[tb]
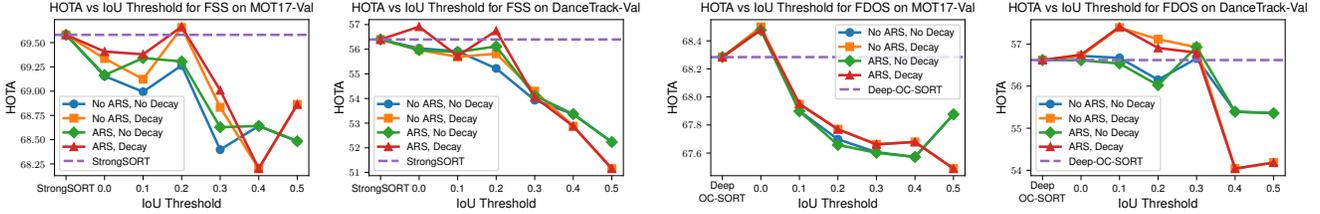

    \begin{raggedleft}
    \resizebox{0.5\textwidth}{!}{
      \input{assets/overall_FSS_MOT17.pgf}
      \input{assets/overall_FSS_DANCE.pgf}
      }
      \resizebox{0.5\textwidth}{!}{
      \input{assets/overall_FDOS_MOT17}
      \input{assets/overall_FDOS_DANCE}
      }
    \end{raggedleft}
    
  \caption{HOTA scores on MOT17 and DanceTrack Validation sets for all parameters that are introduced}
  \label{fig:overall_results}
\end{figure*}

To mitigate the problems caused by missing detections and multiple detections per tracklet, thresholding depending on aspect ratio similarity is introduced. As introduced in the complete IoU loss paper \cite{ciou}, the similarity of the aspect ratios of two bounding boxes is calculated as:

\begin{equation}
    \label{eq:V}
    V = 1 - \frac{4}{\pi^2} \cdot (\arctan\left(\frac{w_1}{h_1}\right) - \arctan\left(\frac{w_2}{h_2}\right))^2
\end{equation}

If the aspect ratio similarity between a detection and its candidate is lower than a threshold. This match is said to be suspicious and the detection is marked as risky to match without features.

Since the significance of aspect ratio similarity depends on IoU, we also adopted the equation \ref{eq:alpha} from complete IoU.

\begin{equation}
    \label{eq:alpha}
    \alpha = \frac{V}{(1 - IoU) + V}
\end{equation}

This calculation's purpose is to adapt the threshold depending on the overlap. For example in Figure \ref{fig:alphavsV}, if the threshold for $\alpha$ ($\theta_\alpha$) is set to 0.6, the effective thresholds for the aspect ratio similarity will be the blue dashed lines. When IoU is 0 or 0.2 the threshold will be 1 meaning that it is marked as risky no matter how similar aspect ratios are. Thus, an implicit lower threshold for IoU is also made.

\subsection{Feature Decay}

As mentioned before, the feature of a tracklet will be updated if the associated detection has features extracted; otherwise, it will remain the same. The feature of a tracklet can mean differently depending on the implementation. If the tracker uses a feature bank mechanism as in DeepSORT \cite{deepsort}, the only side-effect of the mechanism would be the low frequency of adding new features to the bank which can be considered as sampling from the environment with higher intervals. Recent trackers \cite{strongsort, botsort, deepocsort}, generally keep features as an Exponential Moving Average (EMA) of the previous features frame by frame. This is achieved by giving the current feature vector ($e^{t-1}$) a high weight ($\alpha$) and giving the new-coming feature ($f^t$) a low weight as in equation \ref{eq:ema}. This way, the significance of a feature in the average exponentially decays as frames pass. 

\begin{equation}
    \label{eq:ema}
    e^t = \alpha \cdot e^{t-1} + (1-\alpha) \cdot f^t
\end{equation}

In the proposed mechanism, since the features are not updated when matching is done without features, the significance of the earlier features does not decay. To prevent this, we propose to imitate this decaying effect by taking the exponent of $\alpha$ by the number of frames since the last feature extraction from the particular tracklet. This way, the significance of the feature in the average exponentially decays as frames pass.

\begin{equation}
    \label{eq:ema2}
        \begin{cases}
            \alpha^\prime \leftarrow \alpha^\prime \cdot \alpha & \text{if no new feature} \\
            e^t = \alpha^\prime e^{t-1} + (1-\alpha^\prime) f^t; \alpha^\prime \leftarrow \alpha & \text{if new feature}
        \end{cases}
\end{equation}

\section{Experiments}

\subsection{Experimental Setup}

\begin{figure*}[tb]
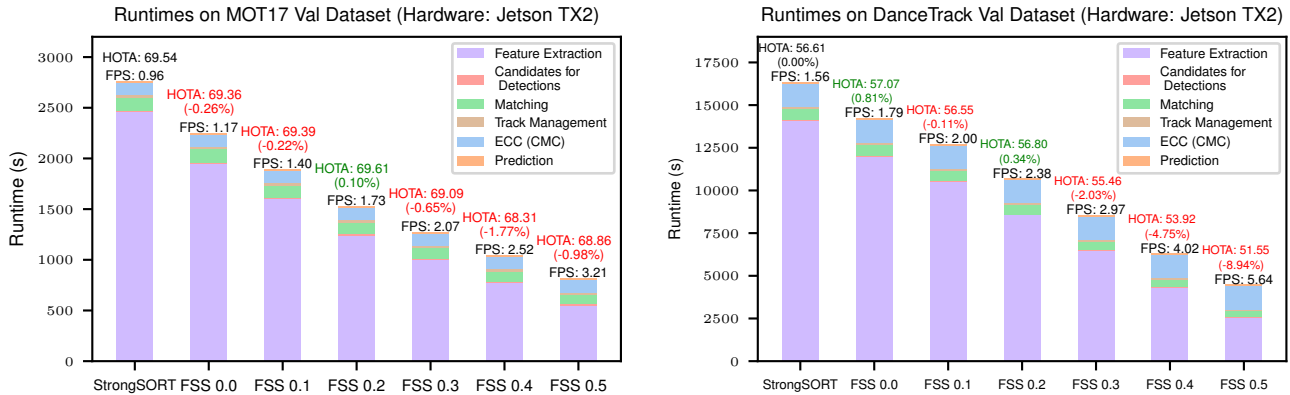

  \begin{raggedleft}
      \input{assets/StrongSort_MOT17_TX2.tex}
      \input{assets/StrongSort_DANCE_TX2.tex}
  \end{raggedleft}
  \caption{Bar plot of share of runtime for different IoU thresholds alongside the change in FPS and HOTA on MOT17 and DanceTrack}
  \label{fig:runtime}
\end{figure*}

We have implemented the mechanism on StrongSORT \cite{strongsort} and Deep OC-SORT \cite{deepocsort}, naming them as Fast-StrongSORT (FSS) and Fast-Deep-OC-SORT (FDOS). We kept all the parameters and the steps of the original methods as they are. Deep-OC-SORT divides the detections into low and high-confidence detections and uses them in the matching phase differently as in ByteTrack \cite{bytetrack}. The mechanism is applied only to the high confidence detections. Both StrongSORT and Deep OC-SORT uses YOLOX \cite{yolox2021} as the object detector, and ResNest50 \cite{resnest} is trained using BoT \cite{bot-train} and SBS \cite{fastreid} methodologies as the feature extractor. StrongSORT provides pre-detected detections and pre-extracted features for MOT17 and MOT20 datasets. We obtained detections for the DanceTrack dataset with the weights shared by OC-SORT. The StrongSORT is made real-time to perform runtime experiments. The runtime experiments are performed on an NVIDIA Jetson TX2 with  8GB of RAM and 256 CUDA cores. 

To make our results comparable with the original methods, we used the pre-extracted features and detections from the original StrongSORT for the experiments that do not include runtime measurements.

The main goal of the experiments was to find out how much gain in FPS can be achieved without losing any HOTA \cite{hota}  score. The change in other metrics such as AssA \cite{hota} and IDF1 \cite{idf1} that are present in the papers of original methods will also be examined.

\subsection{The Best Configuration}
\label{subsec:best_config}

Figure \ref{fig:overall_results} shows the HOTA scores for different combinations of parameters that are being introduced by our mechanism. In general, smaller and higher IoU thresholds result in a decrease in HOTA, and the highest score is attained near 0.2. At first glance, this may seem unintuitive since decreasing the IoU threshold increases the number of detections with features extracted. However, the possible candidates for the detection also increase simultaneously, which enforces the negative effect of missing detections and multiple detections per tracklet (see \ref{subsec:aspect_ratio_similarity}). On the other hand, as the IoU threshold increases, some other tracklets with significant overlap will be ignored. These cases will be further explained in \ref{subsec:effectARS}.

Considering FDOS on MOT17-Val, we inspected the cause of the decrease in the HOTA score on different sequences in MOT17-Val and obtained the results in \ref{tab:FDOS-MOT17-seq}. Two sequences, namely MOT17-02 and MOT17-10, are the main causes of this decrease. Only a single tracklet is responsible for the decrease in both sequences. Thus, for now, we are ignoring this drop.

\begin{table}[ht]
\centering
\caption{Performance comparison of Deep-OC-SORT and FDOS 0.2 on MOT17 sequences.}
\label{tab:FDOS-MOT17-seq}
\begin{tabular}{lcc}
\toprule
Sequence & Deep-OC-SORT & FDOS 0.2 \\
\midrule
MOT17-02 & 47.387 & \textcolor{red}{43.973 (\%-7.20)} \\
MOT17-04 & 78.741 & 78.681 (\% -0.08) \\
MOT17-05 & 59.958 & 60.73 (\%1.29) \\
MOT17-09 & 64.197 & 64.281 (\%0.13) \\
MOT17-10 & 57.766 & \textcolor{red}{56.391 (\%-2.38)} \\
MOT17-11 & 69.108 & 69.091 (\% -0.02) \\
MOT17-13 & 65.194 & 65.67 (\%0.73) \\
\midrule
Overall & 68.284 & 67.769 (\%-0.75) \\
\bottomrule
\end{tabular}
\end{table}

It is also observed that the ARS thresholding does not contribute at all to FDOS while it is beneficial on FSS. The reason lies in the matching strategies of the original methods. Since StrongSORT matches only the features and disregards IoU in the first stage of the matching, it is vital to perform thresholding depending on the shape of the bounding boxes. On the other hand, Deep-OC-SORT uses both IoU and the feature distances in a single step. Considering the bounding boxes whose shapes are different also have less overlap, the Deep-OC-SORT makes this threshold implicitly in its matching phase. In the end, it has no negative effect but may have positive effect to do ARS threshold in any case.


In conclusion, it is undoubtedly beneficial to keep ARS thresholding and feature decay. Hence, we will include them in the following experiments. Throughout our analysis, Symbols FSS and FDOS followed by $\theta_{IoU}$ will be denoting Fast-StrongSORT and Fast-Deep-OC-SORT with ARS and feature decay enabled where the $\theta_ {IoU}$ value is the IoU threshold.

\subsection{Effect of ARS Thresholding}
\label{subsec:effectARS}

\begin{figure}[tb]
    \centering
    
  \includegraphics[width=0.17\textwidth]{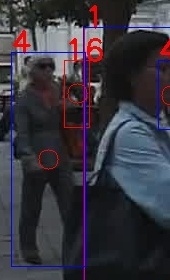}
  \includegraphics[width=0.17\textwidth]{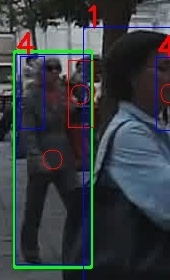}
  \includegraphics[width=0.17\textwidth]{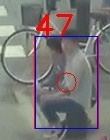}
  \includegraphics[width=0.17\textwidth]{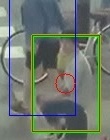}
  \caption{Some cases where erroneous candidacy of a tracklet is caught and thresholded by aspect ratio similarity. \textcolor{blue}{Blue boxes} are detections, \textcolor{red}{red boxes} are tracklets and the \textcolor{green}{green boxes} are the tracklets that were decided to be candidates for multiple or wrong detections, but avoided by ARS thresholding.}
  \label{fig:ARScases}
\end{figure}

As discussed in \ref{subsec:aspect_ratio_similarity}, a tracklet being a candidate for multiple detections is a problem introduced by the base mechanism. In that case, the detection feature will have 0 cosine distance to more than one tracklet, which can result in erroneous matching results. Additionally, even if the matching is not disrupted, the features of some tracklets are updated with a wrong detection, which can have long-term negative effects. An example for this case is the first row of figure \ref{fig:ARScases}, a small detection on the top left of the detection with id 4 has appeared and the tracklet 4 turned out to be a candidate for two detections. Since the aspect ratio similarity between the newly appeared small detection and the tracklet 4 is less than the threshold, this candidacy is avoided. Another problem was the sensitivity against missing detections. The second row of the Figure \ref{fig:ARScases} is an example of such a case. Here, two people are sitting on a bank. Since one of them is occluded, its detection was not obtained. As he stands up, the detection of the person in the front is not received coincidentally. Thus, the ID and the features of the person in the front are copied to the person on the back, and the person in the tracklet has to be given a new ID.

\subsection{Effect of Feature Decay}
\label{subsec:effectfeatdec}

\begin{figure*}[tb]
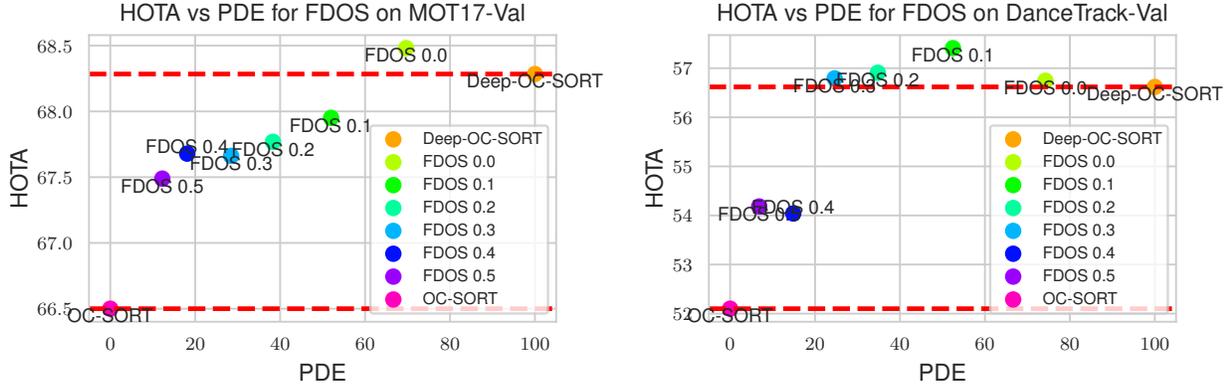

  \begin{center}
      \resizebox{0.96\textwidth}{!}{
      \input{assets/PDE_HOTA_FDOS_MOT17.pgf}
      \input{assets/PDE_HOTA_FDOS_DANCE.pgf}
      }
  \end{center}
  \caption{Percentage of Detections with Extraction vs HOTA score for different IoU thresholds of FDOS}
  \label{fig:runtime_fdos}
\end{figure*}

\begin{figure}[tb]
    \centering
  \includegraphics[width=0.12\textwidth]{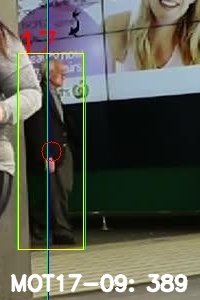}
  \includegraphics[width=0.12\textwidth]{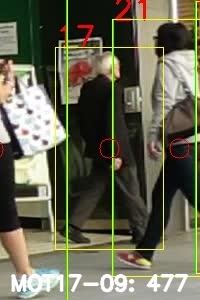}
  \includegraphics[width=0.12\textwidth]{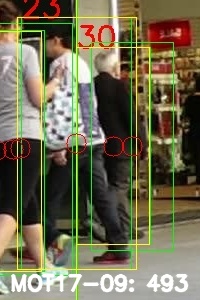}
  \includegraphics[width=0.12\textwidth]{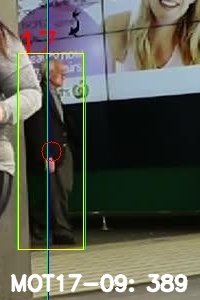}
  \includegraphics[width=0.12\textwidth]{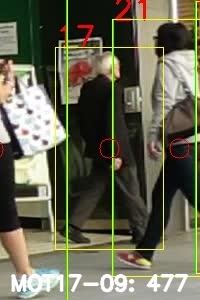}
  \includegraphics[width=0.12\textwidth]{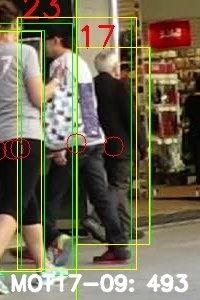}
  \caption{The first row is the output of FSS 0.2 without feature decay; the second row is the output with feature decay. \textcolor{blue}{Blue boxes} are non-risky (without feature) detections, \textcolor{yellow}{yellow boxes} are risky detections and the \textcolor{green}{green boxes} are the tracklets.}
  \label{fig:featdecex}
\end{figure}

\begin{table}[ht]
\centering
\caption{PDE vs Accuracy for Different IoU Thresholds}
\label{tab:runtime_fdos}
\resizebox{0.48\textwidth}{!}{
\begin{tabular}{clcccc}
\toprule
& \textbf{Tracker} & \textbf{PDE* $\downarrow$} & \textbf{HOTA $\uparrow$} & \textbf{AssA $\uparrow$} & \textbf{IDF1 $\uparrow$} \\
\midrule
\multirow{7}{*}{\rotatebox[origin=c]{90}{MOT17 Val}} 
& DOS** & 100.0 & 68.28 & 72.47 & 81.15 \\
& FDOS 0.0 & 69.66 & \textcolor{green}{\textbf{68.48}} & \textcolor{green}{\textbf{72.83}} & \textcolor{green}{\textbf{81.41}} \\
& FDOS 0.1 & 52.00 & 67.95 & 71.69 & 80.42 \\
& FDOS 0.2 & 38.30 & 67.77 & 71.41 & 80.20 \\
& FDOS 0.3 & 28.40 & 67.66 & 71.25 & 80.06 \\
& FDOS 0.4 & 18.11 & 67.68 & 71.26 & 80.10 \\
& FDOS 0.5 & 12.26 & 67.49 & 70.90 & 80.14 \\
\midrule
\multirow{7}{*}{\rotatebox[origin=c]{90}{DanceTrack Val}} 
& DOS** & 100.0 & 56.62 & 41.35 & 57.20 \\
& FDOS 0.0 & 74.24 & \textcolor{green}{\textbf{56.74}} & \textcolor{green}{\textbf{41.53}} & \textcolor{green}{\textbf{57.42}} \\
& FDOS 0.1 & 52.47 & \textcolor{green}{\textbf{57.41}} & \textcolor{green}{\textbf{42.46}} & \textcolor{green}{\textbf{57.91}} \\
& FDOS 0.2 & 34.78 & \textcolor{green}{\textbf{56.91}} & \textcolor{green}{\textbf{41.67}} & \textcolor{green}{\textbf{57.01}} \\
& FDOS 0.3 & 24.62 & \textcolor{green}{\textbf{56.79}} & \textcolor{green}{\textbf{41.56}} & \textcolor{green}{\textbf{56.53}} \\
& FDOS 0.4 & 14.86 & 54.04 & 37.64 & 53.98 \\
& FDOS 0.5 & 6.84 & 54.18 & 37.92 & 54.67 \\
\bottomrule
\end{tabular}
}
\footnotesize
*PDE is the abbreviation of Percentage of Detections with Extraction.
**DOS: Deep OC-SORT
\end{table}

\begin{figure*}
    \centering
    \sffamily
    \setlength\tabcolsep{4pt} 
    \begin{tabular}{p{2pt} c c c c}
    & Intermittent Detections & Poor Localization & Deformation* & Locality of Features \\
    & and Occlusion &  &  & \\
    \rotatebox[origin=l]{90}{\phantom{a}\phantom{a}StrongSORT} & \includegraphics[width=0.10\textwidth]{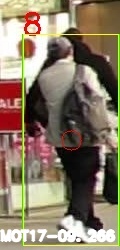}
    \includegraphics[width=0.10\textwidth]{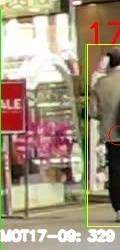} &
    \includegraphics[width=0.10\textwidth]{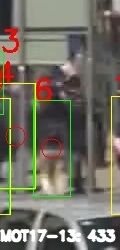} 
    \includegraphics[width=0.10\textwidth]{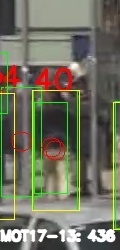} &
    \includegraphics[width=0.10\textwidth]{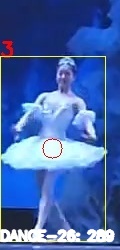} 
    \includegraphics[width=0.10\textwidth]{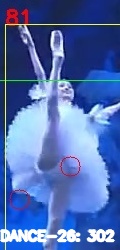} &
    \includegraphics[width=0.10\textwidth]{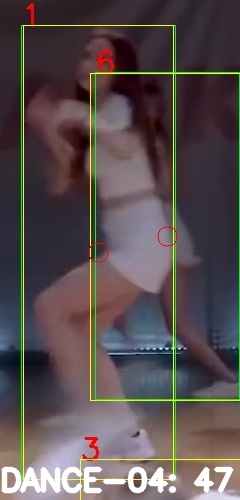} 
    \includegraphics[width=0.10\textwidth]{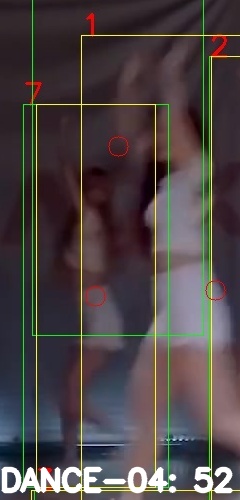} \\

    \rotatebox[origin=l]{90}{\phantom{a}\phantom{a}\phantom{a}\phantom{a}\phantom{a}FSS 0.2} & \includegraphics[width=0.10\textwidth]{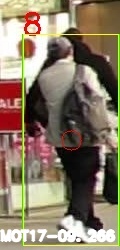}
    \includegraphics[width=0.10\textwidth]{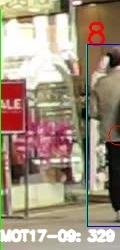} &
    \includegraphics[width=0.10\textwidth]{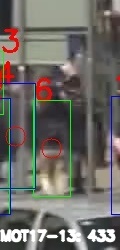}
    \includegraphics[width=0.10\textwidth]{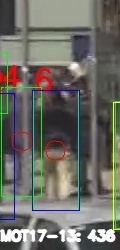} &
    \includegraphics[width=0.10\textwidth]{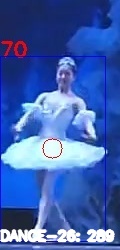} 
    \includegraphics[width=0.10\textwidth]{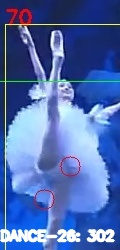} &
    \includegraphics[width=0.10\textwidth]{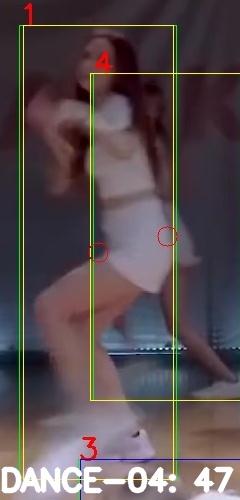} 
    \includegraphics[width=0.10\textwidth]{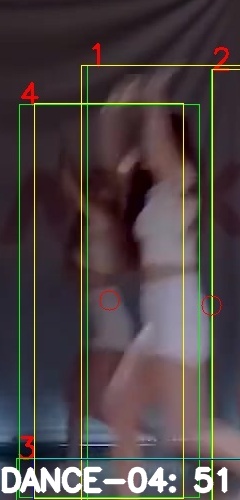} \\
    &\multicolumn{4}{l}{*Outputs for deformation case ($3^{rd}$ column) is obtained without ARS threshollding} \\
    \end{tabular}
    \caption{Example scenarios where our mechanism gives robustness to the original method. The \textcolor{green}{green boxes} are the bounding boxes of the tracklets. The \textcolor{blue}{blue boxes} are the detections that are matched without features. The \textcolor{yellow}{yellow boxes} are the detections that are matched without features.}
    \label{fig:robustness}
    \end{figure*}

Figure \ref{fig:featdecex} shows two example outputs of FSS 0.2 with ARS, one with feature decay and one without it. The person with ID 17 enters into scene in frame 387. His features are extracted in 387-388-389. Between 389 and 477, his detections are non-risky and the features cannot be updated. Between 477 and 493 the detections are always either risky or missing, meaning that the extraction decisions and alphas of EMA are the same for both cases. The effect of feature decay can be seen clearly from this scene. In the first case, the features from frame 389 are preserved and they are given a high weight during the update in frame 477. After a period of occlusion between frames 477 and 493, the main contribution of the feature vector was coming from frame 389. As discussed in \ref{sec:intro}, one of the advantages of feature matching over IoU matching is its robustness against long-term occlusions. Thus, it is intuitive to match features just before the occlusion period with the features right after the occlusion, which is the aim of feature decay.

\subsection{Runtime Analysis}
\label{subsec:runtime}

The bar chart in Figure \ref{fig:runtime} and Table \ref{tab:runtime} show the FPS and HOTA changes in addition to the share of time spent for different IoU thresholds. As stated in section \ref{sec:intro}, feature extraction is responsible for a large portion of the time spent. Our mechanism does reduce the total time spent on the feature extraction phase and increases the FPS as intended. The IoU threshold of 0.2 seems to be the highest threshold that increases the HOTA score and it increases FPS by 80\% on MOT17-Val and 52\% on DanceTrack-Val. Moreover, it is possible to increase FPS by 234\% on MOT17-Val taking into account a 0.98\% decrease in HOTA, which corresponds to tripling the speed. 

The phases other than feature extraction take relatively little time and are not affected significantly by our mechanism. From this point, it can be deduced that the improvement in running time can be expressed as the percentage of detections whose features are extracted. With this new criterion, evaluation can be carried out independently of the working environment and implementation details. This metric is named as PDE.

The Figure \ref{fig:runtime_fdos} and Table \ref{tab:runtime_fdos} present the correlation between HOTA score and PDE for different values of IoU threshold for FDOS. Deep-OC-SORT extends OC-SORT by incorporating feature extraction and camera motion compensation while retaining the same Kalman Filter. The only difference lies in the addition of an embedding cost to the cost matrix. Therefore, the comparison of FDOS with OC-SORT is an admissible way of assessing the extent to which the IoU threshold can be expanded while still maintaining the benefits of feature matching. From the Figure \ref{fig:runtime_fdos}, it is observed that it is possible to increase the IoU threshold up to 0.3 which corresponds to extracting features from 24\% of the detections without sacrificing any accuracy on DanceTrack-Val. Moreover, extracting features from only 6.84\% of the detections still gives +2 HOTA scores compared to the OC-SORT. On MOT17-val, the only threshold that gives better results than the original method appeared to be 0.0 which provides a 25.76\% percent reduction in feature extraction while the threshold of 0.2 which is decided to be the best configuration happened to reduce the accuracy. This reduce is inspected in \ref{subsec:best_config}

\subsection{Test Results}

\begin{table}[ht]
\centering
\caption{Results on Test Sets}
\label{tab:test_results}
\resizebox{0.48\textwidth}{!}{
\begin{tabular}{clcccc}
\toprule
& \textbf{Tracker} & \textbf{PDE* $\downarrow$} & \textbf{HOTA $\uparrow$} & \textbf{AssA $\uparrow$} & \textbf{IDF1 $\uparrow$} \\
\midrule
\multirow{3}{*}{\rotatebox[origin=c]{90}{MOT17}} 
& OC-SORT & 0 & 61.7 & 62.0 & 76.2 \\
& FSS 0.2 & 42.7 & 62.7 & 62.3 & 77.5 \\
& StrongSORT& 100 & \textbf{63.5}  & \textbf{63.7} & \textbf{78.5} \\
\midrule
\multirow{3}{*}{\rotatebox[origin=c]{90}{MOT20}} 
&OC-SORT & 0 & 60.5  & 60.8  & 74.4 \\
&FSS 0.2 & 74.2 & 61.2 & \textcolor{green}{\textbf{72.7}} & 75.4 \\
&StrongSORT& 100 & \textbf{61.5} & 63.2 & \textbf{75.9} \\
\midrule
\multirow{3}{*}{\rotatebox[origin=c]{90}{DANCE}} 
&OC-SORT & 0 & 55.1  & 38.0  & 54.2 \\
&FSS 0.2 & 56.1 & \textcolor{green}{\textbf{55.9}} & \textcolor{green}{\textbf{38.8}} & 54.6 \\
&StrongSORT& 100 & 55.6 & 38.6 & \textbf{55.2} \\
\bottomrule
\multicolumn{6}{c}{*PDE is the abbreviation of Percentage of Detections with Extraction.} \\

\end{tabular}
}
\end{table}

The results on test datasets are provided in Table \ref{tab:test_results}. On DanceTrack dataset FSS 0.2 surpasses StrongSORT. On MOT20 and MOT17 datasets the number of feature extractions is reduced as much as possible with a tiny decrease in the HOTA score. In general, results verify our experiments on validation datasets and demonstrate the power of the proposed mechanism in detecting the only necessary cases for feature extraction.

\section{Conclusion}

Feature extraction is the main cause of overhead yet an essential part of the MOT methods. In this paper, we proposed a mechanism to select the detections that require appearance features to be matched on-the-fly and extract features only from those detections. We observed that it is possible to increase FPS by 80\% and 52\% without sacrificing any accuracy on MOT17 and DanceTrack datasets. Besides improvements in runtime, our mechanism increases accuracy by avoiding confusion during the feature-matching phase in some cases. The question of "When is it really necessary to use appearance features" is critical for further improvements in Multiple Object Tracking.

\newpage

{\small
\bibliographystyle{ieee_fullname}
\bibliography{egbib}
}

\end{document}